\documentclass[runningheads]{llncs}
\usepackage{graphicx}

\usepackage{tikz}
\usepackage{comment}
\usepackage{amsmath,amssymb} 
\usepackage{color}

\usepackage[ruled]{algorithm2e}
\usepackage[accsupp]{axessibility}  

\usepackage{marvosym}
\usepackage{ifsym}

\usepackage{multirow}
\usepackage{diagbox}

\newcommand*\samethanks[1][\value{footnote}]{\footnotemark[#1]}

\def\ie{\emph{i.e.}}

\def\wrt{w.r.t.~}

\newcommand{\figref}[1]{Fig.~\ref{#1}}
\newcommand{\tabref}[1]{Table~\ref{#1}}

\newcommand{\secref}[1]{Sec.~\ref{#1}}
\newcommand{\algref}[1]{Alg.~\ref{#1}}

\newcommand{\addFig}[1]{\includegraphics[width=0.32\linewidth]{figs/#1}}

\begin{document}
\pagestyle{headings}
\mainmatter
\def\ECCVSubNumber{6054}  

\title{ECO-TR: Efficient Correspondences Finding Via Coarse-to-Fine Refinement} 

\titlerunning{ECO-TR}

\author{Dongli Tan\inst{1,3}\thanks{Authors contributed equally.}
\and
Jiang-Jiang Liu\inst{2,3}\samethanks 
\and
Xingyu Chen\inst{3} \and
Chao Chen\inst{3} \and
Ruixin Zhang\inst{3} \and
Yunhang Shen\inst{3} \and
Shouhong Ding\inst{3} \and
Rongrong Ji\inst{1,4} \Letter 
} 
\authorrunning{D. Tan, J-J. Liu, X. Chen, et al.}
%
\institute{Media Analytics and Computing Lab, School of Informatics, Xiamen University \and
TMCC, CS, Nankai University \and 
Youtu Lab, Tencent Technology (Shanghai) Co.,Ltd \and
Institute of Artificial Intelligence, Xiamen University
\\
\email{\{dltan921,j04.liu\}@gmail.com, harleychen@tencent.com, chenchao.tencent@gmail.com, ruixinzhang@tencent.com, shenyunhang01@gmail.com, ericshding@tencent.com, rrji@xmu.edu.cn }
}
\maketitle

\begin{abstract}
Modeling sparse and dense image matching 
within a unified functional correspondence model has recently attracted increasing research interest.
However, existing efforts mainly focus on  
improving matching accuracy while ignoring its efficiency,
which is crucial for real-world applications.
In this paper, 
we propose an efficient structure named Efficient Correspondence Transformer ($\textbf{ECO-TR}$) by finding correspondences in a coarse-to-fine manner, which significantly improves the efficiency of functional correspondence model.
To achieve this, 
multiple transformer blocks are stage-wisely connected 
to gradually refine the predicted coordinates upon 
a shared multi-scale feature extraction network.
Given a pair of images and for arbitrary query coordinates,
all the correspondences are predicted within a single feed-forward pass. 
We further propose an adaptive query-clustering strategy and an uncertainty-based outlier detection module to cooperate with the proposed framework for faster and better predictions.
Experiments on various sparse and dense 
matching tasks demonstrate the superiority of our method 
in both efficiency and effectiveness against existing 
state-of-the-arts.
Project page: \url{https://dltan7.github.io/ecotr/}.

\end{abstract}

\section{Introduction}
As a fundamental research direction in 
computer vision, finding the correspondences 
among pairs of images has been widely utilized 
in plenty of down-stream tasks, including 
optical flow estimation~\cite{liu2010sift,2015flownet,weinzaepfel2013deepflow}, visual localization~\cite{toft2018semantic,sarlin2019coarse,sattler2012improving},
camera position calibration \cite{jin2021image,svarm2016city}, 
3D reconstruction \cite{cheng2014fast,fan2019performance}, and visual tracking \cite{rosten2005fusing}.
Given a pair of images, 
according to how the queries and correspondences are 
determined, 
the applications mentioned above can be generally 
categorized into sparse matching and dense matching.
The former focuses on two sets of keypoints being sparsely and respectively extracted from both 
images and matched to 
minimize a pre-defined alignment error \cite{lowe2004distinctive,sattler2012improving,jin2021image}; 
the latter treats all pixels in the first image as  
queries which are densely mapped to the other image 
for correspondences \cite{liu2010sift,sun2014quantitative,zhou2017unsupervised,ummenhofer2017demon}.

The above two kinds of applications were studied independently for a long time, 
and various optimizations were designed separately.
Recently, COTR~\cite{jiang2021cotr} claims that these two applications can be naturally modeled within a unified framework since the only difference between the sparse and dense matching is the number of points to query.
It proposes to recursively apply a  transformer-based~\cite{carion2020end,vaswani2017attention,dosovitskiy2020image} model at multiple scales in a gradually zooming-in manner to obtain accurate correspondences.
Though impressive performance has been achieved, 
its complex off-line pipeline and slow inference speed seriously limit its practicality in real-world applications.

\begin{figure}[tp]
    \centering
    \includegraphics[width=1.\linewidth]{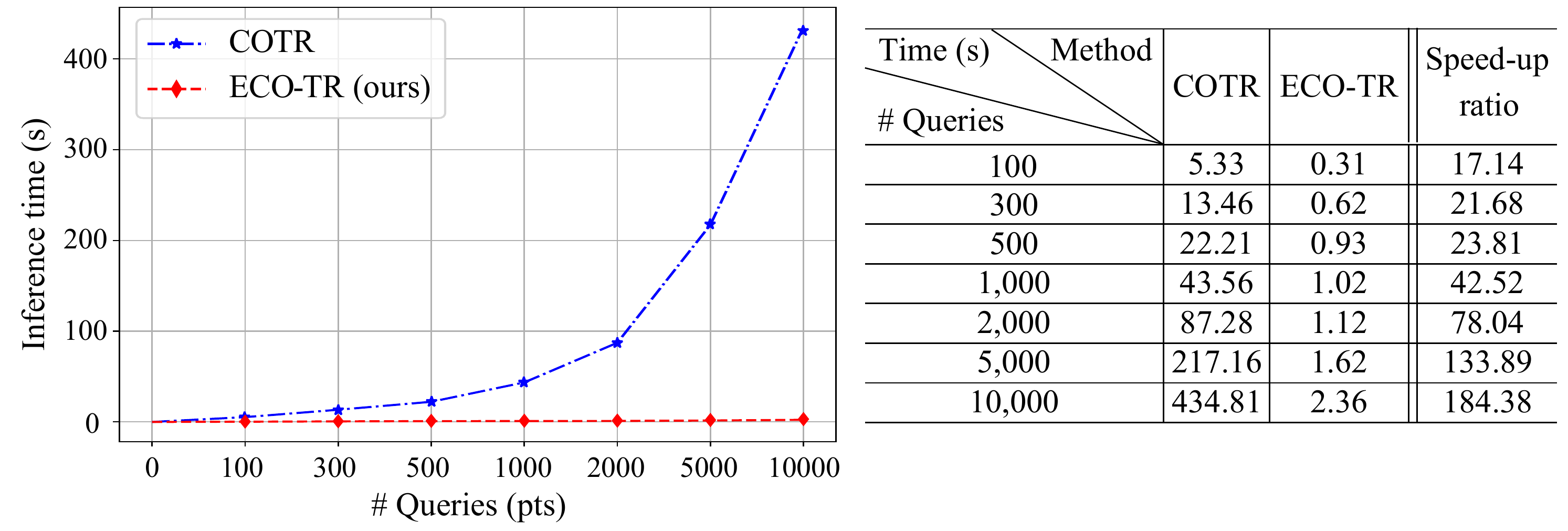}
	\caption{ 
	Comparison of the inference time between the proposed ECO-TR and 
	COTR \cite{jiang2021cotr}. The query numbers are set from 100 to 10,000. 
	As we can see, the time-consuming of COTR increases linearly as the number of points increases, while our method basically does not change.
	}\label{fig:teaser}
\end{figure}

We argue that there are three main reasons leading to the unsatisfactory COTR. The first is 
the recursive zoom-in refinement framework, which must re-extract the corresponding features in the next local patch matching. 
In the case of many queries, these features are likely to overlap, which means plenty of repeated and redundant calculations.
The second is switching the role of the queries and correspondences to 
filter out the mismatched queries, which double the overall computation.
The third is that the staged training strategy leads to unstable training convergence which needs to be carefully fine-tuned.

Instead of sacrificing speed for performance, 
in this work, we present an efficient correspondence transformer network (ECO-TR),
showing that both efficiency and effectiveness can be 
achieved within a single feed-forward pass.
Specifically, we propose to complete the coarse-to-fine refinement
process of the found correspondences in a stage-by-stage manner.
Our framework consists of a bottom-up 
convolutional neural network (CNN) for multi-scale feature
extraction and several top-down transformer blocks corresponding 
to different matching accuracies.
During the coarse-to-fine refinement process, 
rather than cropping image patches of different positions and sizes 
according to the coarsely predicted coordinates and 
recursively re-feeding them into CNN to 
obtain the corresponding feature maps, 
we obtain the multi-scale feature maps \emph{w.r.t.} 
the input image at one time by taking advantages of 
the pyramid and translation invariance nature of modern CNNs, 
and directly crop on the collected feature maps.
The proposed feature-level cropping method can effectively avoid
repeated calculations.
To a certain extent, the inference speed of the model does not increase linearly with the increase of query points.

To further improve the efficiency of our framework, an Adaptive Query-Clustering (AQC) module is proposed to gather similar queries into a cluster, which speeds up the inference.
Moreover, 
we propose an uncertainty module to estimate the confidence of the predicted correspondences, 
which achieves good performance on outlier detection nearly for free.
As illustrated in \tabref{fig:teaser}, our approach can process 1000 queries within one second 
on a single NVIDIA Telsa V100 GPU for a pair of images with
size $800\times 800$, which is around \textbf{40 times} faster than COTR under the same conditions. 

To evaluate the performance of the proposed approach, 
we report the results on multiple challenging datasets 
covering both sparse and dense correspondence finding tasks.
Experimental results demonstrate that our method surpasses COTR in performance and speed by a large margin. 
In addition, we conduct extensive ablation experiments to better understand the impact of each component in our framework. 
The contributions are summarized below:
\begin{itemize}
    \item We propose a new coarse-to-fine framework for finding correspondence that can be applied to both sparse and dense matching tasks. Our method can be optimized end-to-end and evaluate an arbitrary number of queries within a single feed-forward. 
    \item We design an adaptive query-clustering strategy and 
    an uncertainty-based outlier filtering module to achieve
    a better balance between efficiency and effectiveness.
    \item Our method significantly outperforms the existing best-performing functional method in speed and still achieves  comparable performance in sparse correspondence tasks and better in dense correspondence tasks.
\end{itemize}

\section{Related Work}

\subsubsection{Sparse methods.} 

The most common paradigm for sparse image matching pipelines consists of three stages: 
keypoint detection, keypoint description, and feature matching. 
In terms of the detection stage, 
a sparse set of repeatable and matchable keypoints are selected by the
detection methods~\cite{rosten2006machine,savinov2017quad,barroso2019key,tyszkiewicz2020disk}, 
which are robust against viewpoint changes and different lighting conditions. 
Then, the keypoints are described by patch-level input or image-level input. 
Patch-based description methods ~\cite{tian2017l2,mishchuk2017working,tian2019sosnet,ebel2019beyond}
take cropped patches as inputs and are usually trained by metric learning. 
Image-based description methods such as ~\cite{dusmanu2019d2,detone2018superpoint,revaud2019r2d2,luo2020aslfeat,tian2020d2d}
take a full image as input and apply fully-convolutional neural networks~\cite{long2015fully} to generate dense descriptors. 
This kind of method usually combines detector and descriptor, which share the same backbone in training and yield better performance on both tasks.

Traditional feature matching methods use Nearest Neighbor (NN) search 
to find potential matches. Recently, many approaches ~\cite{bian2017gms,yi2018learning,zhang2019learning,zhao2019nm,sun2020acne}
filter outliers by heuristics or learned priors. SuperGlue~\cite{sarlin2020superglue} uses 
an attentional graph neural network and optimal transport method to obtain state-of-the-art performance on sparse matching tasks. 
Unlike the method mentioned above, given some keypoints as queries, 
COTR~\cite{jiang2021cotr} refines the matches in the other image recursively by correspondence neural network. 
Following COTR, we design an end-to-end model to accelerate this 
scheme.

\subsubsection{Dense methods.} 
The main purpose of dense matching is to estimate the optical flow. NC-Net~\cite{rocco2018neighbourhood} represents all keypoints and possible correspondences as a 4D correspondence volume restricted to low-resolution images. Sparse NC-Net~\cite{rocco2020efficient} applies sparse correlation layers instead of all possible correspondences to mitigate this restriction, whereby higher resolution images can be tackled. DRC-Net~\cite{li2020dual} reduces the computational cost and promotes performance by using coarse-resolution and fine-resolution feature maps of different layers. GLU-Net~\cite{truong2020glu} finds pixel-wise correspondences by global and local features extracted from images with different resolutions. GOCor~\cite{truong2020gocor} disambiguates features in similar regions via an improved feature correlation layer. PDC-Net~\cite{truong2021learning} excludes incorrect dense matches in occluded and homogeneous regions by estimating an uncertainty map and filtering the inaccurate correspondences. Patch2Pix~\cite{zhou2021patch2pix} replaces pixel-level matches with patch-level match proposals and later refines them by regression layers. LoFTR~\cite{sun2021loftr} establishes accurate semi-dense matches with linear transformers in a coarse-to-fine manner. For COTR, the dense matching result is generated by interpolating sufficient sparse queries' results. Same with COTR, our method can give dense matching results by interpolation, too.

\subsubsection{Functional methods.} 
The functional method in image matching. COTR is the first one that obtains matches by a functional correspondence finding architecture. Given a pair of images and coordinates of one query, COTR regresses the possible match in the other image via a transformer-based correspondence finding network. Each query is processed independently, and dense correspondences are estimated by interpolating sparse correspondences using Delaunay triangulation of the queries. However, being a recursive method, it will be extremely time-consuming when many keypoints are queried. We mitigate this problem in an end-to-end manner, which runs dozens of times faster than COTR and achieves comparable or superior performance. 

\section{Coarse-to-Fine Refinement Network}

This section describes the proposed end-to-end 
framework that can find the correspondences for arbitrary
queries given a pair of images within a single feed-forward pass 
in detail.

\subsection{Overall Pipeline}

\begin{figure}[tp]
    \centering
    \includegraphics[width=1.\linewidth]{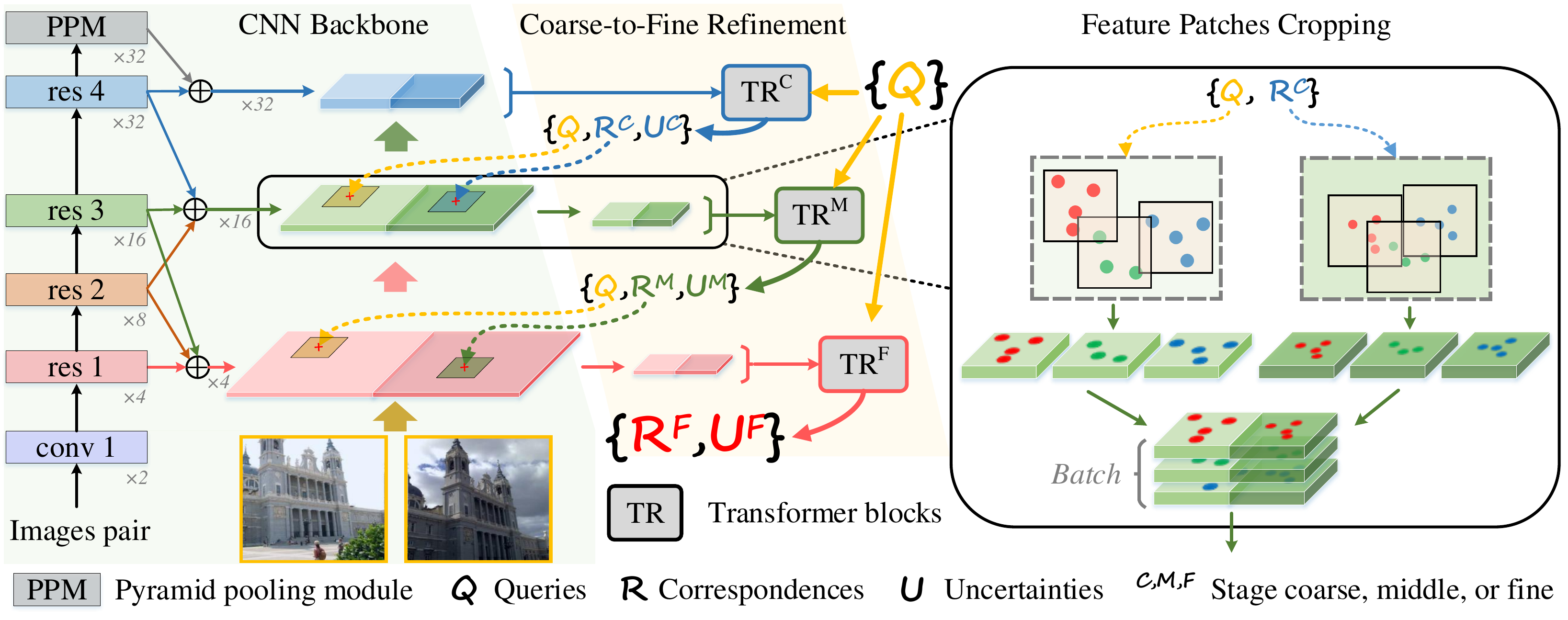}
	\caption{ 
	The pipeline of our proposed framework. 
	It takes a pair of images (bottom-left) 
	and a set of queries (\{\textcolor{orange}{Q}\}) of 
	arbitrary numbers as input and outputs 
	the correspondences (\{\textcolor{red}{$\text{R}^\text{F}$}\})
	and uncertainty scores (\{\textcolor{red}{$\text{U}^\text{F}$}\}), respectively.
	The right part illustrates the feature patches cropping 
	process during each prediction refinement stage.
	}\label{fig:pipeline}
\end{figure}

We show a schematic diagram of the overall pipeline of 
the proposed framework in \figref{fig:pipeline}.
It mainly consists of a bottom-up multi-scale feature extraction  
pathway based on the CNN and a top-down 
coarse-to-fine prediction refinement pathway 
based on the transformer. 
Given a pair of images $I^A$ and $I^B$, 
we first resize them to the same spatial resolution 
($B \times C \times H \times W $, $B$ is the `batch' dimension) and feed them into the CNN backbone to obtain multi-scale features.
After that, the collected multi-scale features
are used along with the input queries to predict 
the correspondences in a coarse-to-fine, gradually refining 
manner in the top-down pathway. 
We also predict an uncertainty score \wrt each correspondence  
representing how confident the network is of its prediction, 
which can be utilized to filter out the outliers nearly for 
free.
Since it could be a bunch of queries to be processed in one
feed-forward, we further introduce an adaptive query-clustering 
strategy to better balance efficiency and effectiveness. 
The following subsections describe the above-mentioned 
components in detail.

\subsection{Efficient Feature Extraction} 
To obtain correspondence locations precisely, 
existing work usually crops image patches around 
potential matching regions 
and iteratively feeds them back into the network 
in a progressively enlarged manner.
The main drawbacks of the aforementioned practice are: 
1) the input image is cropped and resized into patches 
multiple times with different 
zoom-in factors around each query position. 
Each patch generated is then fed into the network, 
which involves many redundant computations.
2) Image patches for each query are cropped and 
processed by the network independently, 
which usually means serial processing and 
inefficient use of computational resources.
We found that the main cause of these two shortcomings 
can be attributed to the setting of cropping patches at 
different spatial levels directly on the image.
Considering the pyramid and translation invariance nature of modern CNNs, 
we propose to alleviate the drawbacks mentioned above by deferring the cropping operation after the feature extraction step. 
Specifically,  
we first obtain the multi-scale feature maps \wrt 
each input image at one time 
and then directly crop on the collected feature maps to 
get feature patches at any position and scale.
We take the ResNet-50\cite{ren2015faster} network as our backbone for multi-scale feature extraction without loss of generality. 
Following the previous success in generating more powerful and 
representative features, 
we attach a pyramid pooling module (PPM) \cite{zhao2017pyramid} to capture more global information at the top of ResNet-50. 
The output of PPM and the side-outputs 
at \texttt{res1-4} stages of the ResNet-50 network 
are collected 
to build a hierarchical multi-scale feature integration structure.
As shown in the left part of \figref{fig:pipeline},
to meet the needs of the subsequent 
top-down pathway which has three refinement stages 
(\ie, coarse, middle, and fine), 
we choose to combine the intermediate outputs of 
\{PPM, \texttt{res4}\}, 
\{\texttt{res2-4}\} and \{\texttt{res1-3}\} stages, respectively. 
The integrated three sets of features 
(denoted as $\{\mathbb{F}^C, \mathbb{F}^M, \mathbb{F}^F \}$) are then resized to 
$1/32$, $1/16$, and $1/4$ spatial resolutions \wrt the 
input stitched images pair, respectively.
%

\renewcommand{\addFig}[1]{\includegraphics[width=0.49\linewidth]{figs/#1}}
\newcommand{\addFigX}[1]{\includegraphics[width=0.98\linewidth]{figs/#1}}
\begin{figure}[tp]
  \centering
  \setlength\tabcolsep{0.5mm}
  \renewcommand\arraystretch{0.6}
  \begin{tabular}{ccc}
  \addFig{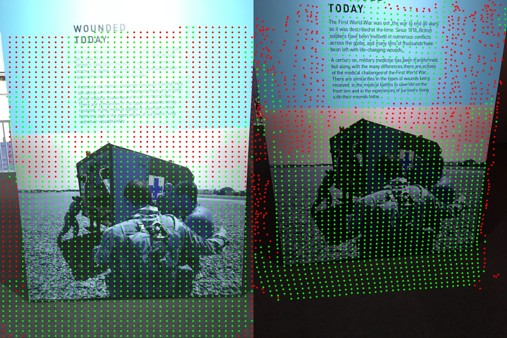} & \addFig{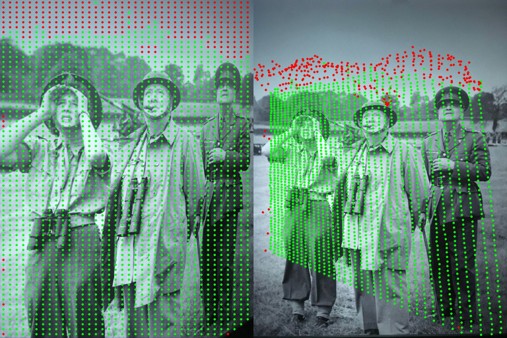} \\
  \end{tabular}
  \caption{Illustration of the uncertainty estimation branch. Green and red points indicate matches with low uncertainties and high  uncertainties, respectively. ECO-TR gives ambiguous predictions in textureless regions and the border area with high uncertainties. }
  \label{fig:uncertainty}
\end{figure}

\subsection{Coarse-to-Fine Prediction Refinement}
The schematic pipeline of the coarse-to-fine prediction refinement process
is shown in the middle part (light orange parallelogram background) 
of \figref{fig:pipeline}. 
Generally speaking, it consists of three successively connected stages: 
coarse, middle, and fine, respectively responsible for 
predicting correspondences with different precision.
Each stage is a transformer building block 
of three encoders and three decoders. 
The coarse stage ($TR^C$) takes a set of queries $\mathbb{Q}$ of arbitrary numbers 
and the entire previously combined features $\mathbb{F}^C$ as input. 
It outputs the coarsely predicted correspondences set $\mathbb{R}^C$ along with their uncertainty scores.
With the guidance of the coordinates 
in $\mathbb{Q}$ and $\mathbb{R}^C$, 
we crop square patches centered at them on the previously collected 
middle-level features $\mathbb{F}^M$ with a fixed 
window size of $w^M$, 
as illustrated by the dashed arrows in
the middle left of \figref{fig:pipeline}. 
The cropped feature patches are then re-arranged into 
a new batch along with the input queries $\mathbb{Q}$
(normalized based on the cropping centers and window sizes) 
being forwarded to the next stage (\ie, the middle stage ($TR^M$)).
The fine stage shares similar procedures 
with the middle stage. 
After the fine stage, 
we obtain the final outputs of the proposed framework:
the finest correspondences $\mathbb{R}^F$ and their uncertainty scores 
$\mathbb{U}^F$. 
%


For each stage, concatenated backbone features are supplemented by 2D linear positional encoding in the sinusoidal format and flattened before being fed into the transformer encoder.
During the decode stage, coordinates of queries with positional encoding attend to the output of the transformer encoder. Here, we disallow self-attention among 
the query points, for queries are independent of each other.
COTR computes the cycle consistency errors and rejects matches whose errors are greater than a specified threshold to filter out uncertain matches, which doubles the computational cost. To further accelerate our framework, we introduce an uncertainty estimation branch. Two FFN branches follow the outputs of the last transformer decoder. One is employed to regress the corresponding relative coordinates of each query, and the other is to predict the uncertainties of these coordinates. Unreliable predictions with high uncertainties will be filtered during the inference stage.

Having predicted matches $\mathbb{R}^i$ and their uncertainties $\mathbb{U}^i$ of level $i$, loss $\mathbb{L}^i$ is calculated by:  
\begin{equation}
\mathbb{L}^{i}=\left\|\mathbb{R}^{i}-\mathbb{R}^{i}_{gt}\right\| \cdot (1-\mathbb{U}^i)+\lambda^{i} \cdot \mathbb{U}^i,   
\end{equation}

where $\mathbb{R}^{i}_{gt}$ is ground truth matches coordinates of queries and $\lambda^{i}$ is the threshold of level $i$, where $i \in \{C, M, F\}$ represents stages coarse, middle, and fine. 
We set $\lambda^{C}=0.1$, $\lambda^{M}=0.05$, $\lambda^{F}=0.01$ during training.

All three stages are supervised during training at the same time. Specifically, the final loss $\mathbb{L}$ is defined as 
\begin{equation}
\mathbb{L}=\mathbb{L}^{C}+\mathbb{L}^{M}+\mathbb{L}^{F}.
\end{equation}
Experiments show that the mid- and fine-level supervision during
training provides predictions for corresponding stages and gives distinctive back-propagation signals to the CNN backbone, which is beneficial to the prediction of coarse-level.
More details are provided in \secref{sec:ablation_studies}.

{
\begin{algorithm}[ht]
    \caption{Adaptive Query-Clustering Algorithm}
    \label{algorithm:AQC}
    \LinesNumbered
    \KwIn {Coordinates of queries $Q$; Matches of $Q$ predicted by previous stage $R$; Iteration number $t$; K-means class number $K_{num}$; Distance threshold $Th$}
    \KwOut {All patch pairs and corresponding matches in these patches}
    \For{i = $1$ to $t$}{
        Divide $Q$ to $K_{num}$ clusters by K-means algorithm, and assign class labels to every pair in $\left(Q,R\right)$ \;
        \For {each class $j$ } {
            Set $\left(Q',R'\right)$ = all pairs labeled $j$ \;
            Set ${c_q}$ = the center coordinates of $Q'$ \;
            Set ${c_r}$ = the center coordinates of $R'$ \;
            \For {each pair $\left(q,r\right)$ in $\left(Q',R'\right)$}{
                \textbf{if} {${\Vert}q-c_q{\Vert}>Th$ or ${\Vert}r-c_r{\Vert}>Th$}{ \\
                    \qquad Set the class label of $\left(q,r\right) = -1$ 
                } 
            }
            Crop patches centered at $c_q$ and $c_r$ and assign pairs labeled $j$ to these patches
        }
        Set $\left(Q,R\right)$ = all pairs labeled -1 \\
    }
    \For {each pair $\left(q,r\right)$ labeled $-1$ in $\left(Q,R\right)$} {
        Crop patches centered at $q$ and $r$, and assign pair $\left(q,r\right)$ to these patches
    }
\end{algorithm}
}

\subsection{Adaptive Query-Clustering}
\label{sec:adaptive_query_clustering}
The transformer structure is capable of processing many queries in one forward propagation. 
To improve efficiency, each patch should contain as many queries as possible. 
A straightforward practice is to 
directly slice the input images pair into two sets of grids 
according to the pre-defined window sizes and strides 
(usually, the stride is set equal to the corresponding window size). 
By densely coupling the patches between these two sets, 
any query-correspondence pair can be assigned to one of the 
patch pairs. 
We denote the above way of point-to-patch assignment as GRID for 
simplicity.
%
%
However, we observe that an inevitable drawback of the query-correspondence 
independent kind of assignment strategies is that some matches will always exist around the patches' borders, 
which usually got sub-optimal matching results. 
We attribute this unsatisfying phenomenon 
to the lack of sufficient contextual information around the border area.

To achieve a better trade-off between efficiency and effectiveness, 
we propose an 
Adaptive Query-Clustering(AQC) algorithm to automatically and dynamically 
assign images patches for all query-correspondence pairs, 
as illustrated in \algref{algorithm:AQC}. 
%
%
To demonstrate the superiority of AQC, we compare it with GRID in \secref{sec:ablation_cluster}. Experiments show that clustering by  AQC gives better performance than GRID.

\subsection{Implementation Details}

We implemented our model in PyTorch~\cite{paszke2019pytorch}. The local feature CNN uses a modified version of ResNet-50 as a backbone without pretraining. For coarse-to-fine refinement modules, we set the crop window size $w^M=17$, $w^F=13$.
For the AQC module, we set $t=1$, $K_{num} = 128 $. The distance threshold $Th$ is set to 0.8 times of the corresponding side of patches during training and 0.6 times during inference. More details can be found in the supplementary material.

\section{Experiments}

We evaluate our method across several datasets. We do not retrain or fine-tune our model on any other dataset for a fair comparison. Experiments are arranged as follows:
\begin{enumerate}
\item Dense matching tasks are evaluated on HPatches~\cite{balntas2017hpatches}, KITTI~\cite{geiger2013vision}, and ETH~\cite{schops2017multi} datasets. Following COTR's evaluation protocol, we evaluate the results of sampled matches and interpolated dense optical flow.
\item 
We evaluate the pose estimation task on the same scene as COTR from Megadepth~\cite{li2018megadepth} dataset for sparse matching.
\item For ablations studies, we evaluate the impact of each proposed contribution using the ETH3D dataset.
\end{enumerate}

\begin{table}[tp]
\center
\caption{\textbf{Quantitative results on HPatches}. Average End Point Error (AEPE) and Percentage of Correct Keypoints (PCK) are reported here. For each method, different thresholds (1px, 3px and 5px) of PCK are used. For a fair comparison of PCK, we report the reproduced results of COTR under the same image size. }
\label{table:DenseHPatches}
\setlength\tabcolsep{1.8mm}
\renewcommand\arraystretch{1.}
\begin{tabular}{l|cccc}
\hline
Method        & AEPE $\downarrow$    & PCK-1px $\uparrow$   & PCK-3px $\uparrow$   & PCK-5px $\uparrow$     \\ \hline
LiteFlowNet~\cite{hui2018liteflownet}   & 118.85               & 13.91                & -                    & 31.64                \\
PWC-Net~\cite{sun2018pwc}       & 96.14                & 13.14                & -                    & 37.14                \\
GLU-Net~\cite{truong2020glu}       & 25.05                & 39.55                & 71.52                & 78.54                \\
GLU-Net+GOCor~\cite{truong2020gocor} & 20.16                & \textbf{41.55}       & -                    & 81.43                \\ 
COTR+Interp (reproduce) ~\cite{jiang2021cotr} & 3.83                 & 36.64                & 76.65                & 87.42                \\ 
ECO-TR+Interp  & \textbf{2.67}        & 40.19       & \textbf{79.89}       & \textbf{90.24}       \\ 
\hline
COTR(reproduce)  ~\cite{jiang2021cotr}        & 3.62                & \textbf{38.72}       & \textbf{80.90}       & \textbf{90.85}       \\
ECO-TR          & \textbf{2.52}        & 38.02                & 79.79                & 90.71                \\

\hline
\end{tabular}
\end{table}

\subsection{Results on HPatches Dataset}
We evaluate ECO-TR on the HPatches dataset for dense matching tasks in the first experiment.
HPatches dataset contains 116 scenes, with 57 scenes changing in viewpoint and 59 scenes changing in lighting conditions. 
Following COTR, we evaluate the dense matching results on viewpoint-changing splits. 
Same with GLU-Net, we resize the reference image during our evaluation, while COTR is evaluated under the original scale in its experiments, which is not comparable in PCK value. Therefore, we reproduce the number of COTR  under fair settings. For each method, we find a maximum of 1,000 matches from each pair. Then, we interpolate correspondences on the Delaunay triangulation map of the queries and get the dense correspondences. The results are reported in \tabref{table:DenseHPatches}. 

For the dense matching task, ECO-TR achieves better performance than COTR under all metrics. For the matching accuracy, COTR is a little better than ECO-TR evaluated by PCK. We attribute this gap to the difference in image resolution. COTR can utilize high-resolution images via four recursive zoom-ins, which is unmanageable for ECO-TR due to its end-to-end architecture. The average endpoint error(AEPE) for ECO-TR is lower than COTR.

\begin{table}[tp]
\center
\caption{\textbf{Quantitative results on KITTI}. Average End Point Error (AEPE) and flow outlier ratio (Fl) on KITTI-2012 and KITTI-2015 are reported below. $\text{COTR}^{\dag}$ means we evaluated it with DenseMatching tools provided by the authors of GLU-Net.}
\label{table:KITTI}
\setlength\tabcolsep{1.8mm}
\renewcommand\arraystretch{1.}
\begin{tabular}{l|cccc}
\hline
\multirow{2}{*}{Method} & \multicolumn{2}{c}{KITTI-2012}                             & \multicolumn{2}{c}{KITTI-2015}                             \\ \cline{2-5} 
& AEPE $\downarrow$ & Fl.{[}\%{]} $\downarrow$ & AEPE $\downarrow$ & Fl.{[}\%{]} $\downarrow$ \\ \hline
LiteFlowNet~\cite{hui2018liteflownet}                              & 4.00                     & 17.47                           & 10.39                    & 28.50                           \\
PWC-Net~\cite{sun2018pwc}                                     & 4.14                     & 20.28                           & 10.35                    & 33.67                           \\
DGC-Net~\cite{melekhov2019dgc}                                     & 8.50                     & 32.28                           & 14.97                    & 50.98                           \\
GLU-Net~\cite{truong2020glu}                                     & 3.34                     & 18.93                           & 9.79                     & 37.52                           \\
RAFT~\cite{teed2020raft}                                        & -                        & -                               & 5.04                     & 17.8                            \\
GLU-Net+GOCor~\cite{truong2020gocor}                               & 2.68                     & 15.43                           & 6.68                     & 27.57                           \\
PDC-Net~\cite{truong2021learning}                                     & 2.08                     & 7.98                           & 5.22                     & 15.13                           \\
$\text{COTR}^{\dag}$ + Interp.~\cite{jiang2021cotr}                                & 1.47                     & 8.79                           & 3.65                     & 13.65                           \\
ECO-TR + Interp.                                & \textbf{1.46}            & \textbf{6.64}                   & \textbf{3.16}       & \textbf{12.10}    \\ \hline
$\text{COTR}^{\dag}$ ~\cite{jiang2021cotr}                                     & 1.15                     & 6.98                            & 2.06                     & 9.14                            \\
ECO-TR                               & \textbf{0.96}            & \textbf{3.77}                   & \textbf{1.40}                     & \textbf{6.39}                           \\
\hline
\end{tabular}
\end{table}

\subsection{Results on KITTI Dataset}
We use the KITTI dataset to evaluate the performance of our method under real road scenes. KITTI2012 dataset contains static scenes only, while the KITTI2015 dataset has more challenging dynamic scenes. Following~\cite{teed2020raft,truong2020gocor,jiang2021cotr}, we use the training split, which has ground truth of camera intrinsics, poses, and depth maps collected by LIDAR. All methods above-mentioned were trained on other datasets and evaluated on this training split. In line with previous works[DGC, GLU, GOC, COTR], We employ the Average End-point Error (AEPE) and percentage of optical flow outliers (Fl) as evaluation metrics. Here, inliers are defined as AEPE$<$3 pixels or $<5\%$. Same with COTR, We sample $40,000$ points for a fair comparison.

As shown in \tabref{table:KITTI}, our method outperforms all others on these two datasets. For example, our method achieves AEPE$=1.09$ and $1.70$ on KITTI-2012 and KITTI-2015, respectively, which is $30\%$ higher than COTR on average. The interpolated results are slightly worse than the sparse results, yet still better than the other dense methods by a large margin, including PDC-Net, which estimates dense correspondence and excludes unreliable matches, too. Qualitative examples on KITTI dataset are illustrated in \figref{fig:optical_flow}.

\renewcommand{\addFig}[1]{\includegraphics[width=0.195\linewidth]{figs/#1}}
\newcommand{\addFigs}[1]{\addFig{kitti_img#1.png} & \addFig{kitti_err_map_cotr_#1.png} & 
\addFig{kitti_err_map_eotr_#1.png} & \addFig{kitti_flow_cotr_#1.png} & \addFig{kitti_flow_eotr_#1.png}}
\begin{figure}[tp]
  \centering
  \small
  \setlength\tabcolsep{0.25mm}
  \renewcommand\arraystretch{1.4}
  \begin{tabular}{ccccc}
  \addFigs{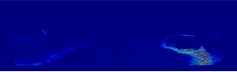} \\
  \addFigs{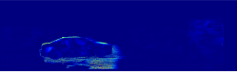} \\
  \addFigs{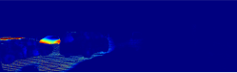} \\
  (a) Input image & (b) COTR & (c) ECO-TR & (d) COTR  & (e) ECO-TR \\
  \end{tabular}
  \caption{ \textbf{Qualitative results on KITTI --} We show the error map (Columns (b, c)) and optical flow (Columns (d, e)) for three pairs from KITTI-2015. ECO-TR provided  clearer outlines of moving objects.}
  \label{fig:optical_flow}
\end{figure}

\subsection{Results on ETH3D Dataset}
ETH3D dataset contains ten image sequences of indoor and outdoor scenes and provides ground truth sparse correspondences under different frame intervals. Following COTR, we report the performance of our method under pairs with seven different intervals, from 3 to 15, respectively. 
The results in \tabref{table:ETH3D} show that our proposal outperforms other competitors under all rates, especially when matching pairs with large geometric transformations, {\textit{i}.\textit{e}.} pairs with a higher rate. 

\begin{table}[tp]
\center
\caption{\textbf{Results on ETH3D.} We evaluated our method over pairs of ETH3D images sampled from different frame intervals. Average End Point Error (AEPE) are reported here. Lower AEPE is better.}
\label{table:ETH3D}
\setlength\tabcolsep{1.1mm}
\renewcommand\arraystretch{1.}
\begin{tabular}{l|ccccccc}
\hline
\multirow{2}{*}{Method} & \multicolumn{7}{c}{AEPE $\downarrow$} \\ 
\cline{2-8} 
& rate=3 & rate=5 & rate=7 & rate=9 & rate=11 & rate=13 & rate=15 \\ \hline
LiteFlowNet~\cite{hui2018liteflownet}                                 & 1.66                       & 2.58                       & 6.05                       & 12.95                      & 29.67   & 52.41   & 74.96   \\
PWC-Net~\cite{sun2018pwc}                                     & 1.75                       & 2.10                       & 3.21                       & 5.59                       & 14.35   & 27.49   & 43.41   \\
DGC-Net~\cite{melekhov2019dgc}                                     & 2.49                       & 3.28                       & 4.18                       & 5.35                       & 6.78    & 9.02    & 12.23   \\
GLU-Net~\cite{truong2020glu}                                     & 1.98                       & 2.54                       & 3.49                       & 4.24                       & 5.61    & 7.55    & 10.78   \\
COTR+Interp.~\cite{jiang2021cotr}                             & 1.71                       & 1.92                       & 2.16                       & 2.47                       & 2.85    & 3.23    & 3.76    \\
ECO-TR+Interp.         & \textbf{1.52}   & \textbf{1.70}     & \textbf{1.87}       & \textbf{2.06}       &\textbf{2.21}       &\textbf{2.44}         &\textbf{2.69}       \\\hline 
COTR~\cite{jiang2021cotr}                          & 1.66   & 1.82     & 1.97       & 2.13       & 2.27        & 2.41        & 2.61    \\
ECO-TR                  & \textbf{1.48}   & \textbf{1.61}     & \textbf{1.72}       & \textbf{1.81}       &\textbf{1.89}       & \textbf{1.97}        & \textbf{2.06}           \\
\hline 
\end{tabular}
\end{table}

\subsection{Results on Megadepth Dataset}

\renewcommand{\addFig}[1]{\includegraphics[width=0.49\linewidth]{figs/#1}}
\begin{figure}[tp]
  {\centering
  \setlength\tabcolsep{0.3mm}
  \renewcommand\arraystretch{1.0}
  \begin{tabular}{ccc}
    \addFig{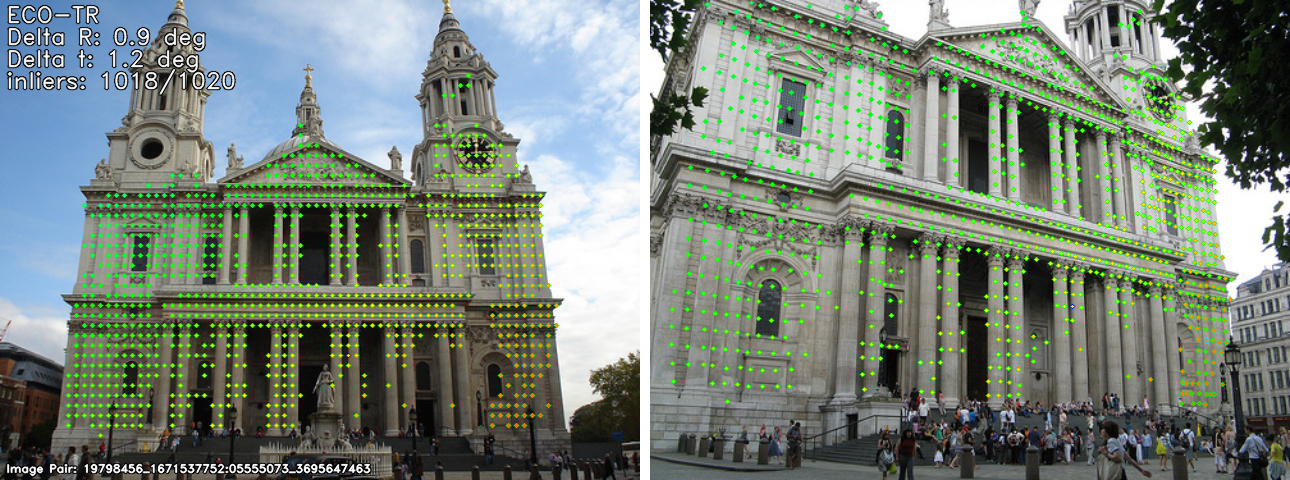} & \addFig{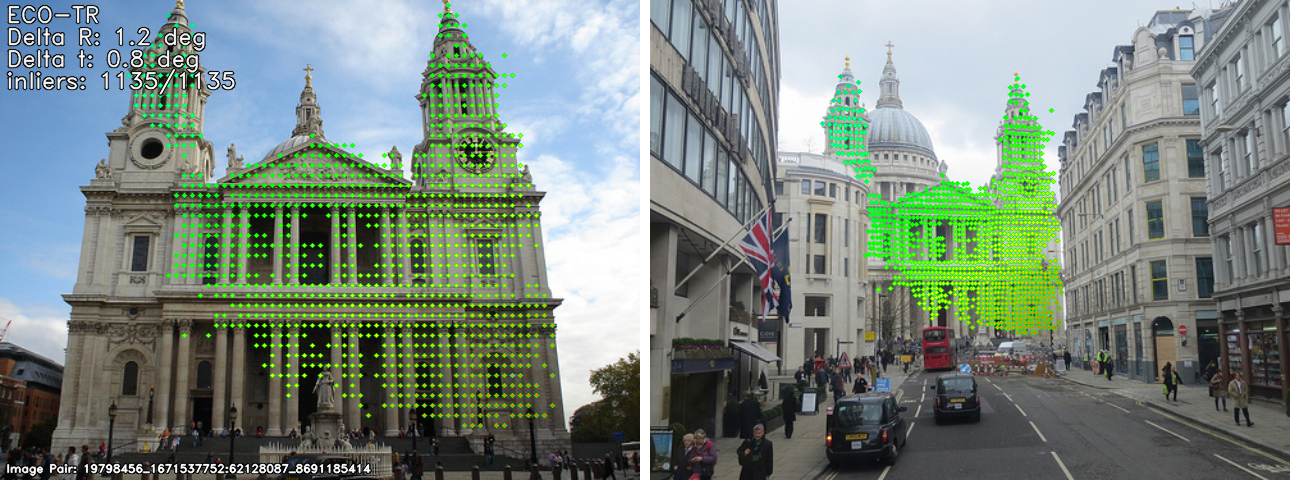} \\
  \end{tabular}
  }
  \caption{\textbf{Qualitative results on MegaDepth dataset}. We set queries on left images and obtain matches in right images. We estimate the relative pose between image pairs and the angular errors
  in rotation and translation are reported in the upper-left corner. The number of inliers evaluated by epipolar distance is shown as well. }
  \label{fig:example_yfcc}
\end{figure}

MegaDepth~\cite{li2018megadepth} images show extreme viewpoint and appearance variations. The poses of images are generated via structure-from-motion and multi-view stereo (MVS) methods, which can be used as ground truth during evaluation. We choose St. Paul's Cathedral as our test scene. We sample 900 pairs of images that have commonly visible regions. Mean average accuracy(mAA) at a {$5$\textdegree} and {$10$\textdegree} error threshold are reported here, where the error is defined as the maximum of angular error in rotation and translation. 
For COTR, we follow the strategy used in its paper and evaluate the performance under different numbers of matches.
For ECO-TR, we estimate the scale of buildings in pairs first. We sample sparse points in one image as queries and predict their correspondences by coarse-stage ECO-TR. Then, we crop original images and obtain patches that share regions of two images. We resize cropped patches and feed them to the model again, and take random points in one image as queries and find reliable matches with low uncertainty in the other image. To further improve performance, a cycle consistency check is applied here. To compare the performance under the same number of matches, we drop some matches randomly. For a fair comparison, other settings except the matching method are fixed for two methods. 
The results in ~\tabref{table:MegaDepth} show that ECO-TR gives a comparable performance, while our pipeline is significantly faster than COTR. Qualitative examples of MegaDepth are illustrated in \figref{fig:example_yfcc}.
%
%

%


\begin{table}[tp]
\begin{center}
\small
\caption{\textbf{Quantitative results on MegaDepth.} We evaluated our method against COTR with different numbers of predicted matches. Mean average accuracy(mAA) at a {$5$\textdegree}  and {$10$\textdegree} error threshold are reported here.}
\label{table:MegaDepth}
\setlength\tabcolsep{0.3mm}
\renewcommand\arraystretch{1.0}
\begin{tabular}{l|cc|cc|cc|cc|cc}
\hline
\multirow{2}{*}{\diagbox[dir=NW]{Method}{\#Matches}} & \multicolumn{2}{c|}{N=2048} & \multicolumn{2}{c|}{N=1024} & 
\multicolumn{2}{c|}{N=512} & \multicolumn{2}{c|}{N=300} & \multicolumn{2}{c}{N=100} \\ \cline{2-11}
& @5 & @10 & @5 & @10 & @5 & @10 & @5 & @10 & @5 & @10 \\ \hline
COTR & 0.443 & 0.660 & 0.448 & \textbf{0.665} & 0.434 & 0.650 & \textbf{0.434} & \textbf{0.654} & 0.410 & 0.626 \\ 
ECO-TR & \textbf{0.453} & \textbf{0.661} & \textbf{0.452} & 0.664 & \textbf{0.447} & \textbf{0.656} & 0.430 & 0.652 & \textbf{0.418} & \textbf{0.636} \\ \hline
\end{tabular}
\end{center}
\end{table}

\subsection{Ablation Studies}
\label{sec:ablation_studies}
In this section, we will conduct several ablation experiments on ETH3D dataset to discuss the efficiency and effectiveness of our method. More ablations on KITTI dataset are provided in the supplementary material.
\subsubsection{Analysis of inference time.}

\tabref{table:ablation_components} reports the time cost of each component of ECO-TR. 
\tabref{table:ECOTR_COTR} further compares 
the runtimes of the corresponding components between ECO-TR and COTR
with similar GPU memory costs (about 8192MB).
As can be seen, all components in ECO-TR are more efficient than COTR's,
where the end-to-end framework (pre- and post-process in an end-to-end manner) contributes most to the efficiency.

\begin{table}[tp]
\caption{Detailed inference time (sec.) of each component.}
\centering
\setlength\tabcolsep{2.mm}
\renewcommand\arraystretch{1.}
\begin{tabular}{c||c|c|c|c|c}
\hline
\#points  & pre- and post-process & backbone & {$TR^C$} & {$TR^M$} & {$TR^F$} \\ \hline
0.1k  &   0.036 & 0.064 & 0.012 & 0.120 & 0.081 \\ \hline
 10k  &   0.037 & 0.062 & 0.026 & 0.480 & 1.740 \\ \hline
\end{tabular}
\label{table:ablation_components}
\end{table}

\subsubsection{Analysis of multistage zoom-ins.}
First, we analyze the effect of multistage zoom-ins architecture. As shown in ~\tabref{table:ablation_on_ETH3D}, we evaluate the result of ECO-TR without middle- and fine-stage inference ($E_{C}$).
It leads to substantially worse results. Adding middle-stage inference benefits the results($E_{CM}$) but is still less effective than three stages version($E_{CMF}$). 
We can see that the design of three-stage refinement is essential for good performance.
Furthermore, instead of training with the supervision of all three branches, we detach the middle-stage and fine-stage branches during training($E_{C'}$). The result shows that it leads to worse results, which indicates that deeply supervised models give more distinctive features which yield better performance. 


\begin{table}[tp]
\caption{Detailed comparison of inference time (sec.) with COTR.}
\centering
\small
\setlength\tabcolsep{1.2mm}
\renewcommand\arraystretch{0.95}
\begin{tabular}{c|c||c|c|c|c}
\hline
Method & \#points & {backbone}&{transformer}&{pre- and post-process} & sum \\ \hline
COTR & 0.1k &  0.67 & 3.74 & 1.03  & 5.44  \\ 
ECO-TR & 0.1k & 0.06 & 0.21 & 0.04  & 0.31 \\ \hline
COTR & 10k & 92.55  & 60.71 & 280.27 & 433.53  \\ 
ECO-TR & 10k & 0.06 & 2.24 & 0.05 & 2.35   \\ \hline
\end{tabular}

\label{table:ECOTR_COTR}
\end{table}

\subsubsection{Analysis of clustering method.}
\label{sec:ablation_cluster}
We test the performance of our pipeline with different clustering methods mentioned in \secref{sec:adaptive_query_clustering}. GRID and AQC are evaluated under the same distance threshold $Th$ for a fair comparison. The results of AQC and GRID clustering are provided in $E_{AQC}$ and $E_{GRID}$ in ~\tabref{table:ablation_on_ETH3D}, respectively.
The result shows that our Adaptive Query-Clustering yields better performance than GRID clustering. 
The gap between the two strategies gradually increases as the difficulty of test pairs increases.


\subsubsection{Analysis of transformer type.}
We replace the full attention transformer block in our middle- and fine-stage model with the linear substitution~\cite{katharopoulos2020transformers} used in LoFTR, and the corresponding results are shown in $E_{linear}$. Compared with full attention result in $E_{fully}$, the AEPE of pairs with rate=3 increases by 0.02 and pairs with rate=3,5 increase by 0.01, while still better than other methods in \tabref{table:ETH3D} by a large margin. Furthermore, the average inference 
time of ECO-TR is reduced by 20 percent when the linear transformer is applied, but this generally leads to a slight degradation in performance. It shows our pipeline has the potential to be further accelerated at a small cost. 


\subsubsection{Analysis of outlier filtering method.}
We compare the effectiveness of the uncertainty-based outlier filtering algorithm in \tabref{table:ablation_on_ETH3D}. We run ECO-TR with different filtering strategies. $E_{cyc}$ employs cycle consistency check as a filter, and $E_{unc}$ employs uncertainty estimation as a filter. The result shows that filtering by uncertainty estimation gives better performance than filtering by cycle consistency check method. Additionally, $E_{cyc+unc}$ employs uncertainty estimation and cycle consistency checks together. Results show that by further using these two strategies together, ECO-TR achieves better performance. 


\begin{table}[tp]
\begin{center}
\caption{\textbf{Ablations on ETH3D.} We evaluate the impact of each component of our method over image pairs from the ETH3D dataset. Pairs are sampled from 3 different frame intervals, which indicate varying difficulty levels. Average End Point Error (AEPE) is reported here. Lower AEPE is better.}
\label{table:ablation_on_ETH3D}
\setlength\tabcolsep{0.1mm}
\renewcommand\arraystretch{1.0}
\begin{tabular}{l||c|c|c|c||c|c||c|c||c|c|c}
\hline
AEPE $\downarrow$ & ~~$E_{C}$~~ & ~~$E_{C'}$~~  & ~$E_{CM}$~ & $E_{CMF}$ & 
$E_{AQC}$ & $E_{GRID}$ & $E_{fully}$ & $E_{linear}$ & $E_{cyc}$ & $E_{unc}$ & $E_{cyc+unc}$ \\ \hline
%
rate=3  & 5.21 & 5.63 & 2.47 & 1.53 & 1.53 & 1.64 & 1.53 & 1.55 & 1.53 & 1.48 & 1.48\\
rate=9  & 7.17 & 7.50 & 3.09 & 2.11 & 2.11 & 2.32 & 2.11 & 2.12 & 2.00 & 1.82 & 1.81\\
rate=15 & 9.19 & 9.53 & 3.83 & 2.72 & 2.72 & 3.10 & 2.72 & 2.74 & 2.45 & 2.08 & 2.06\\ \hline
\end{tabular}
\end{center}
\end{table}

\section{Conclusions}
This paper introduces an efficient coarse-to-fine transformer-based network for local feature matching. 
The main improvement is from three sides: 1) We propose an efficient network structure in a coarse-to-fine manner, fully utilizing the information from different layers and can be trained integrally. 2) We design an adaptive query-clustering (AQC) module that gathers similar query points in the same patch and achieves a better balance between efficiency and effectiveness. 
3) An uncertainty-based outlier detection module is proposed to filter out the queries without correspondence. 
Our method significantly improves the speed of functional matching and achieves comparable or better performance both on sparse and dense matching tasks. 

\subsubsection{Limitations}
The main limitation is that the training of ECO-TR requires a large amount of GPU computing resources. In addition, simple interpolation and refinement techniques limit the performance of dense estimates. We leave these for the future work.

\subsubsection{Acknowledgments}
This work was supported by the National Science Fund for Distinguished Young Scholars (No.62025603), the National Natural Science Foundation of China (No. U21B2037, No. 62176222, No. 62176223, No. 62176226, No. 62072386, No. 62072387, No. 62072389, and No. 62002305), Guangdong Basic and Applied Basic Research Foundation(No.2019B1515120049), and the Natural Science Foundation of Fujian Province of China (No.2021J01002).

\clearpage
%
%
\bibliographystyle{splncs04}
\bibliography{egbib}
\end{document}